\documentclass[conference, letterpaper]{IEEEtran}
\IEEEoverridecommandlockouts

\usepackage{cite}
\usepackage{amsthm}
\usepackage{amsmath,amssymb,amsfonts}
\usepackage{algorithm}
\usepackage{algorithmic}
\usepackage{graphicx}
\usepackage{textcomp}
\usepackage{multirow}
\usepackage{multicol}
\usepackage{url}
\usepackage{subfigure}
\usepackage{xcolor}
\usepackage[letterpaper, right=1.7cm, left=1.7cm, top=1.9cm, bottom=3.45cm]{geometry}
\def\BibTeX{{\rm B\kern-.05em{\sc i\kern-.025em b}\kern-.08em
    T\kern-.1667em\lower.7ex\hbox{E}\kern-.125emX}}

\begin{document}
\title{Joint Optimization of Memory and Computing Frequency for Energy-Efficient DNN Inference}
\author{\IEEEauthorblockN{
Yunchu~Han\IEEEauthorrefmark{1},~Zhaojun~Nan\IEEEauthorrefmark{2},~Sheng~Zhou\IEEEauthorrefmark{1},~and~Zhisheng~Niu\IEEEauthorrefmark{1}
}
\IEEEauthorblockA{\IEEEauthorrefmark{1}Beijing National Research Center for Information Science and Technology\\Department of Electronic Engineering, Tsinghua University, Beijing 100084, China}
\IEEEauthorblockA{\IEEEauthorrefmark{2}School of Electronics and Internet of Things\\Chongqing Polytechnic University of Electronic Technology, Chongqing 401331, China
}
\IEEEauthorblockA{Emails: hyc23@mails.tsinghua.edu.cn,~nanzhaojun@cquet.edu.cn,~\{sheng.zhou@,~niuzhs@\}tsinghua.edu.cn}}
\maketitle

\begin{abstract}
Deep neural network (DNN) inference on mobile devices often incurs high latency and energy consumption due to limited computing and memory resources. To enable energy-efficient DNN inference, most existing studies focus on dynamic voltage and frequency scaling (DVFS) for adjusting the computing frequency, while the impact of memory frequency on the inference performance has been greatly overlooked. In this paper, we consider the impact of memory frequency and computing frequency on DNN inference time, and jointly optimize these two frequencies together with communication resources for energy-efficient DNN inference. Based on a realistic inference time model, we formulate an optimization problem to minimize the energy consumption of all mobile devices under the deadline constraint. For local inference, we derive a near-optimal closed-form solution via convex optimization, while an optimal closed-form solution for transmission power is obtained for edge inference with the given bandwidth. Furthermore, we propose a low-complexity heuristic algorithm to effectively solve the overall problem with polynomial time complexity. Simulation results based on measured data show that the proposed near-optimal solution for local inference can achieve optimal performance under strict deadline constraints, with a performance gap of up to $2.5\%$ compared with the optimal solution. Meanwhile, our proposed algorithm significantly reduces the energy consumption of devices by up to $10.4\%$ compared to other methods.
\end{abstract}

\section{Introduction}\label{sec:intro}
The rapid development of deep neural networks (DNNs) has significantly advanced computer vision and artificial intelligence applications \cite{Liu1}, \cite{Cao2}. However, the intensive computation and memory access requirements of DNN inference impose critical challenges for mobile devices with limited energy and computing resources. To enable low-latency and energy-efficient DNN inference, mobile edge computing (MEC) \cite{Mao3} has emerged as a promising technology, allowing computation tasks to be offloaded from mobile devices to nearby edge servers. Furthermore, edge intelligence \cite{Zhou4} has been proposed as a new paradigm that enables real-time and intelligent services at the wireless network edge (e.g., base stations and road side units). However, mobile devices still need to manage their local computing and memory resources efficiently to balance inference latency and energy consumption.

To address this issue, the dynamic voltage and frequency scaling (DVFS) \cite{Han5}, \cite{Rabaey6} technique has been widely studied to achieve a tradeoff between latency and energy consumption by dynamically adjusting the operating frequency of processors. Existing DVFS-based approaches mainly focus on the adjustment of computing frequency (e.g., CPU frequency, GPU frequency) to reduce latency or energy consumption. For example, the authors in \cite{Nan7} design a robust DNN partitioning and resource allocation algorithm to handle uncertain inference time, where DVFS is adopted to adjust the computing frequency for local inference. In addition, DVFS can be combined with the batch processing technique to reduce the energy consumption \cite{Shi8} or increase the throughput \cite{Nabavinejad}.

\begin{figure}[t]
  \centering
  \includegraphics[width=0.35\textwidth]{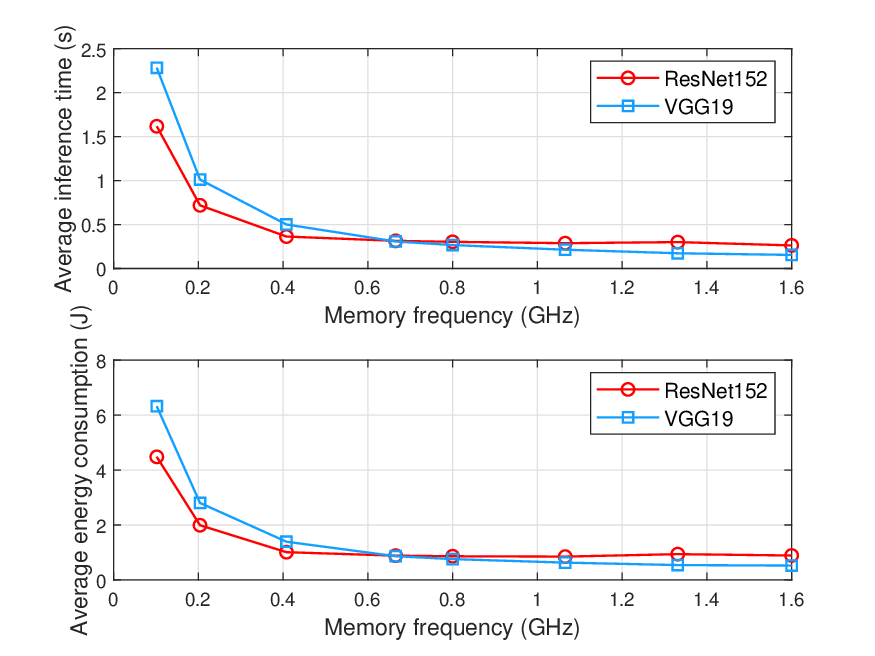}\\
  \caption{Impact of memory frequency scaling on DNN inference time and energy consumption for ResNet152 and VGG19 on Jetson TX1.}
  \label{fig:Intro}
  \vspace{-0.3cm}
\end{figure}
However, recent studies have revealed that memory frequency scaling also plays a significant role in the overall latency, energy and accuracy characteristics of DNN inference, especially for memory-intensive DNN models \cite{Han10}, \cite{Zhang11}, \cite{Wang12}, \cite{Denkinger}. In \cite{Han10}, the impact of joint memory frequency and computing frequency scaling on DNN inference time for edge devices is characterized and analyzed. In \cite{Zhang11}, the impact of memory frequency and computing frequency on energy efficiency of DNN inference is investigated, respectively, and a reinforcement learning algorithm is proposed to optimize these frequencies. In \cite{Wang12}, the authors mainly analyze the impact of memory frequency and computing frequency on kernel-based applications. In \cite{Denkinger}, the impact of memory voltage scaling on accuracy and resilience of DNNs for edge devices is presented. To further investigate the impact of memory frequency scaling on latency and energy consumption of DNN inference, we deploy ResNet152 \cite{HeK} and VGG19 \cite{Simonyan} on Jetson TX1 to evaluate the corresponding performance, as shown in Fig. \ref{fig:Intro}. It is observed that by increasing the memory frequency from $0.1$ GHz to $1.6$ GHz, the average inference time can be reduced by $84\%$ and $93\%$ for ResNet152 and VGG19, respectively. Meanwhile, $80\%$ and $92\%$ reductions can be achieved for the average energy consumption by only adjusting the memory frequency. Therefore, memory frequency scaling can also significantly impact the inference time and energy consumption. Despite these potentials, the joint optimization of memory frequency and computing frequency for energy-efficient DNN inference has not been fully studied. 

In this paper, we consider an edge intelligence system that adopts the joint optimization of memory frequency, computing frequency and communication resources to reduce the energy consumption of mobile devices. Specifically, we derive an optimal solution to the local inference problem in some special cases. Moreover, we analyze the upper bound of the corresponding problem and derive a near-optimal closed-form solution. Meanwhile, the analysis for the edge inference problem provides an optimal closed-form solution for the transmission power, and a heuristic algorithm is proposed to efficiently solve the overall optimization problem. Simulation results show that the proposed near-optimal solution achieves a performance within $2.5\%$ of the optimal result for local inference, and reduces the energy consumption by up to $10.4\%$ compared with other methods, demonstrating the effectiveness of the proposed approach. 

\section{System Overview}\label{sec:system}
\subsection{System Model}
\begin{figure}[t]
  \centering
  \includegraphics[width=0.35\textwidth]{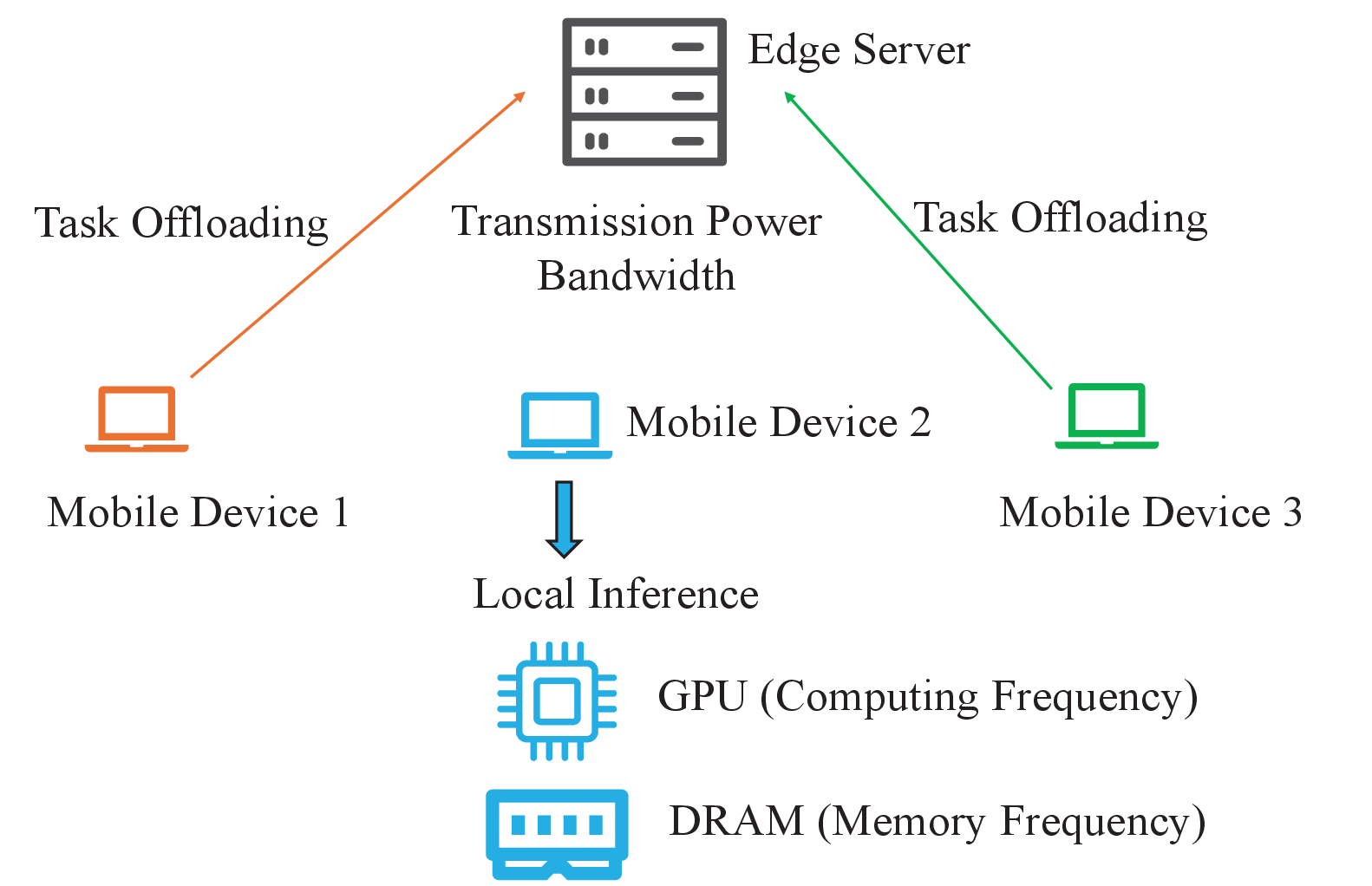}\\
  \caption{{Illustration of an edge intelligence system that jointly optimizes memory and computing frequencies, transmission power and bandwidth.}}
  \label{fig:SystemModel}
  \vspace{-0.3cm}
\end{figure}

As shown in Fig. \ref{fig:SystemModel}, we consider an edge intelligence system consisting of $N$ mobile devices and an edge server. The set of all mobile devices is denoted as $\mathcal{N} \triangleq \{1, 2, \dots, N\}$. The orthogonal frequency division multiple access (OFDMA) technology is adopted to reduce the interference among mobile devices, and we assume that the bandwidth allocated to each device does not overlap. Each device needs to execute DNN inference tasks within the given deadline $D_n$. We consider the binary offloading mode (i.e., $x_n \in \{0, 1\}$) in this work, where mobile device $n$ can either execute local inference (i.e., $x_n = 1$) or edge inference by offloading the inference task (i.e., $x_n = 0$). For local inference, mobile devices can adjust the memory frequency and computing frequency to save the energy consumption of DNN inference with the deadline constraint. For edge inference, mobile devices can change the transmission power to improve the transmission energy while meeting the communication latency constraint. The edge inference time is negligible due to the powerful computing capacity of the MEC server, and the energy consumption of edge inference is ignored here because the MEC server is usually powered by the grid \cite{ref}. In addition, the latency of downloading inference results is ignored due to their small data volume \cite{Nan}.

\subsection{Inference Time and Energy Consumption Model}
We perform real-world experiments to obtain the average inference time under different combinations of memory frequencies and computing frequencies \cite{Han10}, and the impact of memory frequency $f_{n, \mathrm{mem}}$ and computing frequency $f_{n, \mathrm{com}}$ on DNN inference time is formulated as
\begin{equation}\label{eq:myeq1}
    t_n^{\mathrm{loc}} = a_n f_{n, \mathrm{mem}}^{-b_n} + c_n f_{n, \mathrm{com}}^{-d_n}, \forall n \in \mathcal{N},
\end{equation}
The dynamic power consumption of CMOS circuit is denoted by $P = \alpha C V^2 f$, where $\alpha$, $C$, $V$ and $f$ denote the activity factor, the capacity, the supply voltage and the frequency, respectively \cite{Haj}. Since the power consumption consists of the memory and computing power \cite{Guerreiro}, the power consumption of mobile device $n$ for executing inference is written as
\begin{align}\label{eq:myeq2}
    p_n^{\mathrm{loc}} = \kappa_{n, \rm{mem}} f_{n, \rm{mem}}^3 + \kappa_{n, \rm{com}} f_{n, \rm{com}}^3 + \sigma_n, \forall n \in \mathcal{N},
\end{align}
where the details are given in \cite{Han10}. Therefore, the energy consumption of mobile device $n$ for executing inference is
\begin{align}\label{eq:myeq3}
    e_n^{\mathrm{loc}} = p_n^{\mathrm{loc}} t_n^{\mathrm{loc}}, \forall n \in \mathcal{N}.
\end{align}

\subsection{Transmission Time and Energy Consumption Model}
The transmission rate of mobile device $n$ can be written as
\begin{align}\label{eq:myeq4}
    r_n = B_n \log_2 \left( 1 + \frac{p_n h_n}{\sigma^2} \right), \forall n \in \mathcal{N},
\end{align}
where $B_n$, $p_n$, $h_n$ and $\sigma^2$ denote the allocated bandwidth, transmission power, channel gain and noise power, respectively. Then, the transmission latency for task offloading of mobile device $n$ can be written as
\begin{align}\label{eq:myeq5}
    t_n^{\mathrm{tran}} = \frac{s_n}{r_n}, \forall n \in \mathcal{N},
\end{align}
where $s_n$ is the data volume of the computation task. The corresponding energy consumption of transmission is
\begin{align}\label{eq:myeq6}
    e_n^{\mathrm{tran}} = p_n t_n^{\mathrm{tran}}, \forall n \in \mathcal{N}.
\end{align}

\section{Problem Formulation and Solutions}\label{sec:formulation}
We focus on minimizing the sum of the energy consumption of all mobile devices by optimizing the binary offloading decision $x_n$, transmission power $p_n$, bandwidth $B_n$, memory frequency $f_{n, \mathrm{mem}}$ and computing frequency $f_{n, \mathrm{com}}$, while the deadline constraint should be satisfied. The optimization problem can be formulated as
\begin{subequations}\label{eq:Problem1}
\begin{alignat}{2}
& \textbf{P1:} \min _{x_n, p_n, B_n, f_{n, \mathrm{mem}}, f_{n, \mathrm{com}}} \sum_{n \in \mathcal{N}} \left( x_n e_n^{\mathrm{loc}} + \left( 1 - x_n \right) e_n^{\mathrm{tran}} \right) \label{eq:Problem1a} \\
& \ \text {s.t.} \, x_n t_n^{\mathrm{loc}} + \left( 1 - x_n \right) t_n^{\mathrm{tran}} \leq D_n, \forall n \in \mathcal{N}, \label{eq:Problem1b} \\
& \ \quad \ \, f_{n, \min} \leq f_{n, \mathrm{mem}} \leq f_{n, \max}, \forall n \in \mathcal{N}, \label{eq:Problem1c} \\
& \ \quad \ \, F_{n, \min} \leq f_{n, \mathrm{com}} \leq F_{n, \max}, \forall n \in \mathcal{N}, \label{eq:Problem1d} \\
& \ \quad \ \, p_{n, \min} \leq p_n \leq p_{n, \max}, \forall n \in \mathcal{N}, \label{eq:Problem1e} \\
& \ \quad \ \, x_n \in \{0, 1\}, \forall n \in \mathcal{N}, \label{eq:Problem1f} \\
& \ \quad \ \, \sum_{n \in \mathcal{N}} x_n B_n \leq B, \label{eq:Problem1g} \\
& \ \quad \ \, B_n \geq 0, \forall n \in \mathcal{N}, \label{eq:Problem1h}
\end{alignat}
\end{subequations}
where (\ref{eq:Problem1b}) is the deadline constraint, (\ref{eq:Problem1c}) is the memory frequency constraint, (\ref{eq:Problem1d}) is the computing frequency constraint, (\ref{eq:Problem1e}) is the transmission power constraint, (\ref{eq:Problem1f}) is the binary offloading decision constraint, (\ref{eq:Problem1g}) and (\ref{eq:Problem1h}) denote the bandwidth constraint, respectively. Since the variables $x_n$ are binary and coupled in the objective function, Problem P1 is a mixed-integer nonlinear programming (MINLP) problem. To simplify Problem P1, we first focus on the single-user scenario. An optimal solution is solved under specific conditions, while a near-optimal closed-form solution is derived for more general cases. Moreover, for edge inference, we derive an optimal closed-form solution for the transmission power with the given bandwidth. Finally, we propose a heuristic algorithm to solve Problem P1, which greedily searches possible devices to execute edge inference. The detailed analysis and solution are presented as follows.

First, consider a special single-user case of Problem P1, and the corresponding problem is formulated as
\begin{subequations}\label{eq:Problem2}
\begin{alignat}{2}
& \textbf{P2:} \min _{x_n, p_n, B_n, f_{n, \mathrm{mem}}, f_{n, \mathrm{com}}}  x_n e_n^{\mathrm{loc}} + \left( 1 - x_n \right) e_n^{\mathrm{tran}}  \label{eq:Problem2a} \\
& \ \text {s.t.} \, x_n t_n^{\mathrm{loc}} + \left( 1 - x_n \right) t_n^{\mathrm{tran}} \leq D_n, \forall n \in \mathcal{N}, \label{eq:Problem2b} \\
& \ \quad \ \, f_{n, \min} \leq f_{n, \mathrm{mem}} \leq f_{n, \max}, \forall n \in \mathcal{N}, \label{eq:Problem2c} \\
& \ \quad \ \, F_{n, \min} \leq f_{n, \mathrm{com}} \leq F_{n, \max}, \forall n \in \mathcal{N}, \label{eq:Problem2d} \\
& \ \quad \ \, p_{n, \min} \leq p_n \leq p_{n, \max}, \forall n \in \mathcal{N}, \label{eq:Problem2e} \\
& \ \quad \ \, x_n \in \{0, 1\}, \forall n \in \mathcal{N}, \label{eq:Problem2f} \\
& \ \quad \ \, 0 \leq B_n \leq B, \forall n \in \mathcal{N}. \label{eq:Problem2g}
\end{alignat}
\end{subequations}
We first consider the case where mobile device $n$ performs local inference (i.e., $x_n = 1$). Then, we can obtain the following optimization problem:
\begin{subequations}\label{eq:Problem3}
\begin{alignat}{2}
& \textbf{P3:} \min _{f_{n, \mathrm{mem}}, f_{n, \mathrm{com}}} e_n^{\mathrm{loc}}  \label{eq:Problem3a} \\
& \ \text {s.t.} \, a_n f_{n, \mathrm{mem}}^{-b_n} + c_n f_{n, \mathrm{com}}^{-d_n} \leq D_n, \forall n \in \mathcal{N}, \label{eq:Problem3b} \\
& \ \quad \ \, f_{n, \min} \leq f_{n, \mathrm{mem}} \leq f_{n, \max}, \forall n \in \mathcal{N}, \label{eq:Problem3c} \\
& \ \quad \ \, F_{n, \min} \leq f_{n, \mathrm{com}} \leq F_{n, \max}, \forall n \in \mathcal{N}. \label{eq:Problem3d} 
\end{alignat}
\end{subequations}
Notice that the constraints (\ref{eq:Problem3b}), (\ref{eq:Problem3c}) and (\ref{eq:Problem3d}) are convex, but the objective function is not always convex. However, by simple computation, we can prove that the objective function of Problem P3 is convex if $0 \leq b_n \leq 2$ and $0 \leq d_n \leq 2$. In this specific case, Problem P3 is a standard convex optimization problem, and the optimal solutions can be solved by optimization tools (e.g., CVX \cite{Grant}).

Moreover, we try to analyze Problem P3 and give a closed-form solution under a general case. Since the inference time $t_n$ should be no larger than the deadline $D_n$, the upper bound of the objective function (\ref{eq:Problem3a}) is given by replacing $t_n$ with $D_n$. The corresponding optimization problem is formulated as
\begin{subequations}\label{eq:Problem4}
\begin{alignat}{2}
    & \textbf{P4:} \min _{f_{n, \mathrm{mem}}, f_{n, \mathrm{com}}} \left( \kappa_{n, \mathrm{mem}} f_{n, \mathrm{mem}}^3 + \kappa_{n, \mathrm{com}} f_{n, \mathrm{com}}^3 + \sigma_n \right) D_n \label{eq:Problem4a} \\
    & \ \text{s.t.} \, (\mathrm{\ref{eq:Problem3b}}), (\mathrm{\ref{eq:Problem3c}}), (\mathrm{\ref{eq:Problem3d}}), 
\end{alignat}
\end{subequations}
which is a standard convex optimization problem. Based on the Karush-Kuhn-Tucker (KKT) conditions \cite{Boyd}, we can derive the optimal closed-form solutions to Problem P4. The details are given in the following lemma. 

\emph{Lemma 1:} The optimal memory frequency and computing frequency to Problem P4 are given by
\begin{align}
    & f_{n, \mathrm{mem}}^{\ast} = \max(f_{n, \min}, \tilde{f}_{n, \mathrm{mem}}), \label{eq:Singlefm_optimal} \\
    & f_{n, \mathrm{com}}^{\ast} = \max(F_{n, \min}, \tilde{f}_{n, \mathrm{com}}), \label{eq:Singlefc_optimal}
\end{align}
where $\tilde{f}_{n, \mathrm{mem}}$, $\tilde{f}_{n, \mathrm{com}}$ and the Lagrange multiplier $\tilde{\lambda}_n$ should satisfy  
\begin{align}
    & \tilde{f}_{n, \mathrm{mem}} = \left( \frac{\tilde{\lambda}_n a_n b_n}{3 D_n \kappa_{n, \mathrm{mem}}} \right)^{\frac{1}{b_n + 3}}, \label{eq:Singlef_mtilde} \\
    & \tilde{f}_{n, \mathrm{com}} = \left( \frac{\tilde{\lambda}_n c_n d_n}{3 D_n \kappa_{n, \mathrm{com}}} \right)^{\frac{1}{d_n + 3}}, \label{eq:Singlef_ctilde} \\
    & a_n \tilde{f}_{n, \mathrm{mem}}^{-b_n} + c_n \tilde{f}_{n, \mathrm{com}}^{-d_n} = D_n. \label{eq:Singlelambda}
\end{align}

\begin{proof}
    Denote $\lambda_n \geq 0$ as the Lagrange multiplier, and the Lagrange function can be written as
    \begin{align}
        & L(f_{n, \mathrm{mem}}, f_{n, \mathrm{com}}, \lambda_n) = \nonumber \\
        & \left( \kappa_{n, \mathrm{mem}} f_{n, \mathrm{mem}}^3 + \kappa_{n, \mathrm{com}} f_{n, \mathrm{com}}^3 + \sigma_n \right) D_n + \nonumber \\
        & \lambda_n \left( a_n f_{n, \mathrm{mem}}^{-b_n} + c_n f_{n, \mathrm{com}}^{-d_n} - D_n \right).
    \end{align}
    Based on the KKT conditions, we have
    \begin{align}
        & \frac{\partial L}{\partial f_{n, \mathrm{mem}}} \Bigg|_{\tilde{f}_{n, \mathrm{mem}}} = 0, \label{eq:SingleKKT1} \\
        & \frac{\partial L}{\partial f_{n, \mathrm{com}}} \Bigg|_{\tilde{f}_{n, \mathrm{com}}} = 0, \label{eq:SingleKKT2} \\
        & \tilde{\lambda}_n \left( a_n \tilde{f}_{n, \mathrm{mem}}^{-b_n} + c_n \tilde{f}_{n, \mathrm{com}}^{-d_n} - D_n \right) = 0. \label{eq:SingleKKT3}
    \end{align}
    By simple computation, we can get (\ref{eq:Singlef_mtilde}) and (\ref{eq:Singlef_ctilde}). Since the frequency is positive, the Lagrange multiplier should satisfy $\tilde{\lambda}_n > 0$. Combining with (\ref{eq:SingleKKT3}), we can get (\ref{eq:Singlelambda}). Note that the left side of (\ref{eq:Singlelambda}) is monotonically decreasing with the increasing of $\tilde{\lambda}_n$, there exists a unique solution for equation (\ref{eq:Singlelambda}). In addition, $\tilde{f}_{n, \mathrm{mem}}$ and $\tilde{f}_{n, \mathrm{com}}$ should be in the feasible range, which means that $f_{n, \mathrm{mem}}^{\ast}$ ( $f_{n, \mathrm{com}}^{\ast}$ ) should take the larger one between $f_{n, \min}$ ($F_{n, \min}$) and $\tilde{f}_{n, \mathrm{mem}}$ ($\tilde{f}_{n, \mathrm{com}}$). Therefore, the optimal memory frequency and computing frequency are given by (\ref{eq:Singlefm_optimal}) and (\ref{eq:Singlefc_optimal}), respectively. 
\end{proof}

Subsequently, we focus on the optimization of transmission power with the given bandwidth when $x_n = 0$, and the optimization problem is expressed as
\begin{subequations}\label{eq:Problem5}
\begin{alignat}{2}
\textbf{P5:} &\min _{p_n} \frac{p_n s_n}{r_n} \label{eq:Problem5a} \\
& \ \text {s.t.} \, \frac{s_n}{r_n} \leq D_n, \forall n \in \mathcal{N}, \label{eq:Problem5b} \\
& \ \quad \ \, p_{n, \min} \leq p_n \leq p_{n, \max}, \forall n \in \mathcal{N}. \label{eq:Problem5c} 
\end{alignat}
\end{subequations}
Although the constraint (\ref{eq:Problem5b}) is convex, Problem P5 is not a convex optimization problem due to the non-convexity of (\ref{eq:Problem5a}). However, the objective function of Problem P5 is a monotonically increasing function of transmission power $p_n$. Based on this observation, we can give an optimal closed-form solution, and the details are shown as follows.

\emph{Lemma 2:} The optimal transmission power for Problem P5 is given by
\begin{equation}
    p_n^{\ast} = \max(p_{n, \min}, \tilde{p}_n), \label{eq:poptimal}
\end{equation}
where
\begin{equation}
    \tilde{p}_n = \frac{\left( 2^{\frac{s_n}{b_n D_n}}-1 \right) \sigma^2 }{h_n}. \label{eq:ptilde}
\end{equation}

\begin{proof}
    Denote $f(p_n) = \frac{p_n s_n}{r_n}$ as the objective function of Problem P5, where $r_n = B_n \log_2 \left( 1 + \frac{p_n h_n}{\sigma^2} \right)$. Then, we have
    \begin{equation}
        \frac{\partial f}{\partial p_n} = \frac{s_n \log2}{B_n} \frac{\log\left( 1+\frac{p_n h_n}{\sigma^2} \right) - \frac{p_n h_n}{p_n h_n + \sigma^2}}{\log^2 \left( 1 + \frac{p_n h_n}{\sigma^2} \right)}.
    \end{equation}
    Since $\log(\frac{1}{1+x}) \leq \frac{1}{1+x} - 1$, we get $\log\left( 1 + x \right) \geq \frac{x}{1 + x}$. Let $x = \frac{p_n h_n}{\sigma^2}$, we can prove that $\frac{\partial f}{\partial p_n} \geq 0$, which shows that (\ref{eq:Problem5a}) monotonically increases with $p_n$. The optimal value of Problem P5 is achieved at the minimum $p_n$ in the feasible range. The deadline constraint (\ref{eq:Problem5b}) requires that the transmission power should be equal or greater than $\tilde{p}_n$. In addition, $p_n^{\ast}$ should satisfy the constraint (\ref{eq:Problem5c}), taking the larger value between $\tilde{p}_n$ and $p_{n, \min}$. Therefore, $p_n^{\ast}$ is given by (\ref{eq:poptimal}).
\end{proof}

Based on the above analysis, we propose a heuristic algorithm to effectively solve Problem P1. In each iteration, the total bandwidth is equally allocated among all offloading devices, and the optimal transmission power is obtained by solving Problem P5. If the maximum transmission power among these devices satisfies the available power budget $p_{n, \max}$, Algorithm 1 terminates and the corresponding resource allocation is determined. Otherwise, the device requiring the highest transmission power is assigned to perform local inference, and the algorithm is repeated for the remaining devices. The detailed procedure is summarized in Algorithm \ref{alg:1}. Since each iteration involves solving convex optimization problems and the maximum number of iterations is $N$, our proposed algorithm can solve Problem P1 with polynomial computational complexity.
\begin{algorithm}[t]
\renewcommand{\algorithmicrequire}{\textbf{Input:}} 
\renewcommand{\algorithmicensure}{\textbf{Output:}} 
\caption{Heuristic Algorithm for Solving Problem P1}
\label{alg:1}
\begin{algorithmic}[1]
\REQUIRE Total bandwidth $B$, number of devices $N$, deadline $D_n$, channel gain $\{h_j\}_{j=1}^N$, power budget $\{p_{j, \max}\}_{j=1}^N$;
\ENSURE Offloading policy $x_n$, transmission power $p_n$, bandwidth $B_n$, memory frequency $f_{n, \mathrm{mem}}$, computing frequency $f_{n, \mathrm{com}}$;
\STATE Initialize $\mathcal{U} \leftarrow \{1,2,\ldots,N\}$;
\WHILE{true}
    \STATE $N_{\mathrm{act}} \leftarrow |\mathcal{U}|$;
    \STATE Equally allocate bandwidth $B_n = B / N_{\mathrm{act}}$, $\forall n \in \mathcal{U}$;
    \STATE Compute the optimal $p_n^{\ast}$ for each device based on Lemma 2;
    \STATE Find the maximum $p_{\mathrm{th}} = \max_{j \in \mathcal{U}} p_j$ and corresponding device $j^\star$;
    \IF{$p_{\mathrm{th}} \leq p_{n, \max}$}
        \STATE \textbf{break} \quad // All devices satisfy the power budget
    \ELSE
        \STATE Remove device $j^\star$ from $\mathcal{U}$;  \quad // Eliminate the infeasible device
        \STATE Solve the near-optimal $f_{n, \mathrm{mem}}^{\ast}$ and $f_{n, \mathrm{com}}^{\ast}$ based on Lemma 1;  \quad // Execute local inference
    \ENDIF
    \IF{$\mathcal{U} = \emptyset$}
        \STATE \textbf{break} \quad // No feasible offloading solution
    \ENDIF
\ENDWHILE
\STATE \textbf{Return} $\mathcal{U}$, $p_n^{\ast}$, $B_n$, $f_{n, \mathrm{mem}}^{\ast}$, $f_{n, \mathrm{com}}^{\ast}$.
\end{algorithmic}
\end{algorithm}

\begin{table*}[t]
    \caption{\label{tab:II}\textsc{Parameters of DNNs and Devices.}}
	\centering
	\begin{tabular}{c | c | c | c | c | c | c | c | c | c | c }
		\hline
		{DNN Model} &{Mobile Device} & $f_{n, \mathrm{mem}}$ (GHz) & $f_{n, \mathrm{com}}$ (GHz) & {$a_n$} & {$b_n$} & {$c_n$} & {$d_n$} & {$\kappa_{n, \mathrm{mem}}$} & {$\kappa_{n, \mathrm{com}}$} & {$\sigma_n$} \\ 
        \hline
		{ResNet152} &\multirow{2}{*}{{Jetson TX1}} &\multirow{2}{*}{[0.102, 1.6]} &\multirow{2}{*}{[0.1536, 0.9984]} &{0.039} &{1.552} &{0.211} &{0.358} &\multirow{2}{*}{{0.152}} &\multirow{2}{*}{{1.115}} &\multirow{2}{*}{{2.097}} \\
        \cline{1-1}\cline{5-8}
        {VGG19} &  &  &  &{0.110} &{1.296} &{0.099} &{0.629} &  &  &  \\
        \hline
	\end{tabular}
    \vspace{-0.2cm}
    \label{tab:mytab1}
\end{table*}
\section{Simulation Results}\label{sec:eval}
We assume that there are $N = 12$ mobile devices in a square area of $400 ~\mathrm{m} \times 400 ~\mathrm{m}$, and the total bandwidth is $B = 20$ MHz. The 3GPP channel model $h_n = 38 + 30 \log_{10} l_n$ \cite{3GPP} is adopted here, where $l_n$ (in meters) and $h_n$ (in dB) are the distance and path loss, respectively. The transmission power range is $[0.1, 1]$ W. ResNet152 and VGG19 are deployed on Jetson TX1 to execute DNN inference on CIFAR-100 dataset \cite{Krizhevsky}. The detailed parameters are shown in Table \ref{tab:mytab1}.

\begin{figure}[t]
  \centering
  \includegraphics[width=0.4\textwidth]{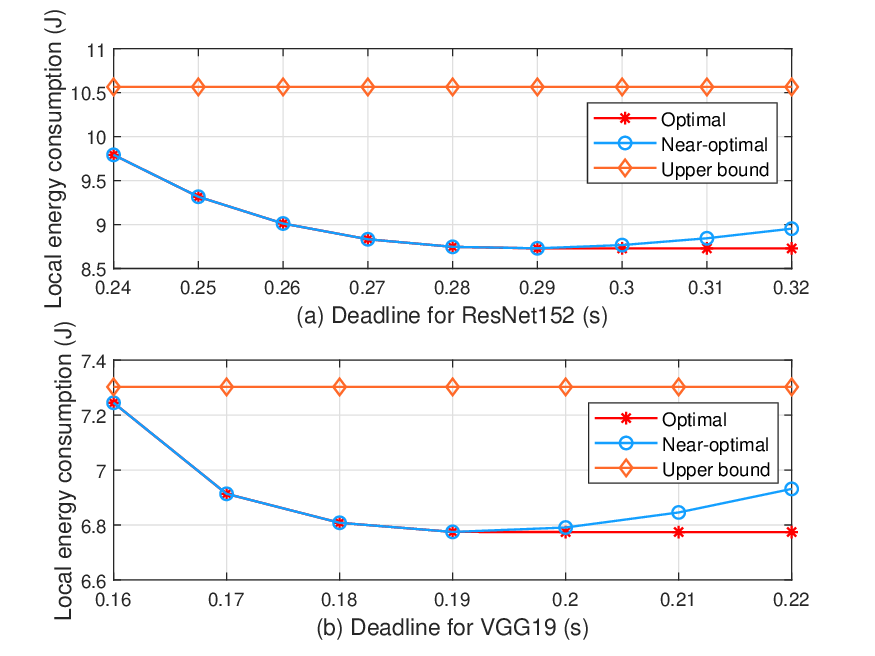}\\
  \caption{Comparison for the local energy consumption of optimal and near-optimal solutions under different deadlines.}
  \label{fig:Energy_Compare}
  \vspace{-0.3cm}
\end{figure}

Fig. \ref{fig:Energy_Compare} presents the total energy consumption of mobile devices for optimal and near-optimal solutions under various deadlines for local inference on Jetson TX1 with ResNet152 and VGG19. The upper bound is given by executing local inference with the maximum memory frequency and computing frequency. When the deadline is relatively small, the near-optimal solution can achieve the optimal performance. In this case, the optimal energy consumption is achieved when the inference time is equal to the deadline. However, as the deadline further increases, the performance gap increases. For ResNet152 with $D_n = 0.32$ s and VGG19 with $D_n = 0.22$ s, the near-optimal solution leads to $2.5\%$ and $2.3\%$ errors, respectively. Therefore, the performance of near-optimal solution is guaranteed when the deadline is relatively small, where we can obtain the optimal performance with lower computational complexity. In addition, the energy consumption of near-optimal solution does not exceed the upper bound under various deadlines, which further shows the effectiveness of our proposed near-optimal solution.

\begin{figure}[t]
  \centering
  \includegraphics[width=0.4\textwidth]{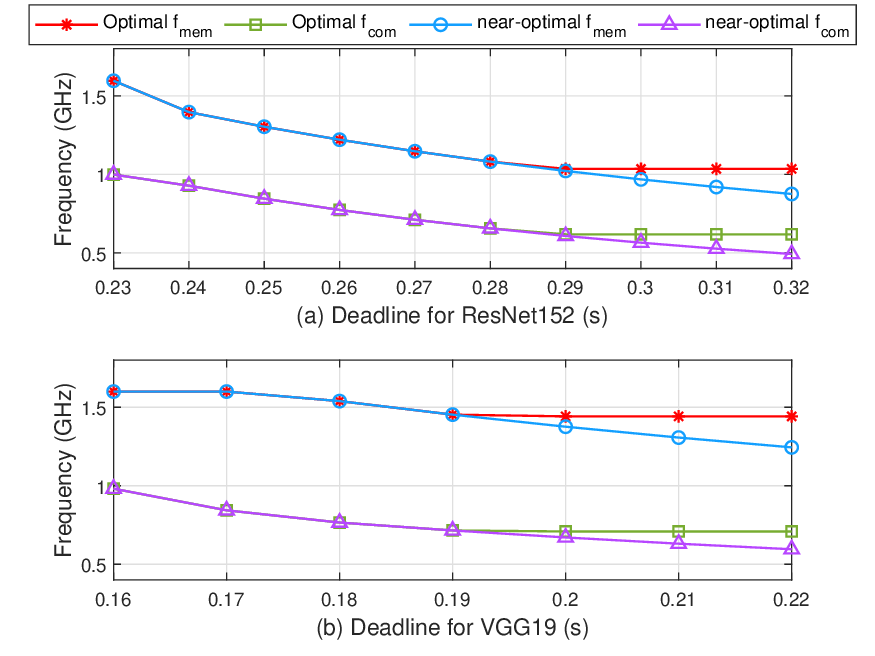}\\
  \caption{Comparison for the memory frequency and computing frequency of optimal and near-optimal solutions under different deadlines.}
  \label{fig:Frequency_Compare}
  \vspace{-0.3cm}
\end{figure}

In Fig. \ref{fig:Frequency_Compare}, we evaluate the corresponding memory frequency and computing frequency for optimal and near-optimal solutions under different deadlines for local inference. It can be observed that the near-optimal memory frequency and computing frequency decrease as the deadline increases, since expressions (\ref{eq:Singlef_mtilde}) and (\ref{eq:Singlef_ctilde}) monotonically decrease with the deadline $D_n$. However, the optimal memory frequency and computing frequency do not decrease but remain constant when the deadline exceeds a threshold, which is consistent with the results shown in Fig. \ref{fig:Energy_Compare}. In this range, the reduction in power consumption caused by the decrease of frequencies can not compensate for the increase of inference time. Specifically, for ResNet152 with $D_n = 0.32$ s and VGG19 with $D_n = 0.22$ s, the gap between optimal and near-optimal memory frequency is $15.4\%$ and $13.7\%$, while the computing frequency gap is $20.3\%$ and $16.0\%$, respectively.

Finally, we evaluate our proposed Algorithm 1 under different bandwidth resources and device numbers. For comparison, we consider the following three benchmarks. 
\begin{itemize}
    \item Random: Instead of removing the device with the maximum transmission power, randomly remove one device from the set $\mathcal{U}$.
    \item Only Compute \cite{Nan24}: Instead of jointly optimizing the two frequencies, only the computing frequency is optimized in Algorithm 1.
    \item No DVFS: Instead of jointly optimizing the two frequencies, the memory frequency and computing frequency remain at the maximum values in Algorithm 1.
\end{itemize}

\begin{figure}[t]
  \centering
  \includegraphics[width=0.38\textwidth]{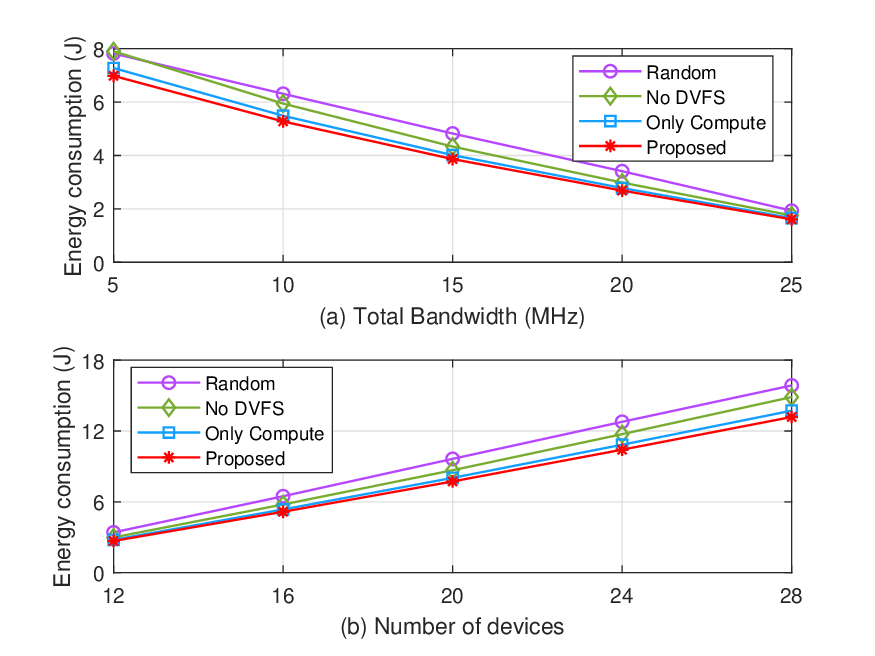}\\
  \caption{Energy consumption under various bandwidth resources and devices.}
  \label{fig:Transmission}
  \vspace{-0.3cm}
\end{figure}

In Fig. \ref{fig:Transmission}, we present the energy consumption for different solving policies. As the total bandwidth increases, more tasks are offloaded from devices to the edge server, which leads to further improvement in energy consumption for all policies. Compared with the Only Compute and No DVFS policies, our proposed Algorithm 1 can achieve average energy savings of $3.6\%$ and $10.4\%$, respectively, demonstrating the advantage of jointly optimizing the memory and computing frequency. Furthermore, the proposed algorithm significantly outperforms the Random policy, since it fails to effectively exploit the channel characteristics. It is also observed that when the number of devices is small or the available bandwidth is sufficiently large, the performance gap among different policies becomes less pronounced.

\section{Conclusion}\label{sec:conclusion}
In this paper, we have studied the joint optimization of memory frequency, computing frequency, bandwidth and transmission power for energy-efficient DNN inference. By integrating analytical modeling for DNN inference time and convex optimization methods, we derive the optimal solution for the local inference problem under specific cases, and a near-optimal closed-form solution is derived for general cases. With the given bandwidth, an optimal closed-form solution for transmission power is derived for the edge inference problem. Finally, a heuristic algorithm is proposed to solve the problem with low computational complexity. Simulation results further verify the characteristics of analytical solutions, where the proposed near-optimal solution can achieve optimal performance in specific cases, and the performance gap with the optimal solution is no more than $2.5\%$. Compared with other methods, our proposed algorithm can substantially reduce the energy consumption of mobile devices.

\section{Acknowledgement}
This work is supported in part by the Open Fund of State Key Laboratory of Intelligent Green Vehicle and Mobility, Tsinghua University, and in part by the State Key Laboratory of Internet of Things for Smart City (University of Macau) Open Research Project under Grant SKL-IoTSC(UM)/ORP04/2026, and in part by the Project of Tsinghua University-Toyota Joint Research Center for AI Technology of Automated Vehicle under Grant TTAD-2025-08.



\end{document}